\pdfoutput=1
\documentclass[11pt]{article}

\usepackage[preprint]{acl}
\usepackage{times}
\usepackage{latexsym}

\usepackage[T1]{fontenc}
\usepackage[utf8]{inputenc}

\usepackage{microtype}

\usepackage{inconsolata}

\usepackage{xcolor}

\usepackage{booktabs} % for using toprule, midrule etc. in tables
\usepackage{listings} % used for examples in conll format
\usepackage{graphicx} % used for adding figures
\usepackage{fancyvrb} % box around verbatim text
\usepackage[colorinlistoftodos]{todonotes} % todo notes
\usepackage{makecell} % to use in tables, for text flow
\usepackage{float} % to force figure placing
\usepackage{graphicx}  % Optional, for better formatting of wide tables
\usepackage{color, colortbl} % For coloring rows
\definecolor{Gray}{gray}{0.9}
\definecolor{Green}{rgb}{0.7, 1, 0.7}
\usepackage{pifont}% http://ctan.org/pkg/pifont
\newcommand{\cmark}{\ding{51}}%
\newcommand{\xmark}{\ding{55}}%


\newcommand{\approach}{\textsc{VerbalizED}}

\title{Evaluating Design Decisions for Dual Encoder-based Entity Disambiguation}

\author{Susanna Rücker \\
  Humboldt-Universität zu Berlin \\
  \texttt{susanna.ruecker@hu-berlin.de} \\\And
  Alan Akbik \\
  Humboldt-Universität zu Berlin \\
  \texttt{alan.akbik@hu-berlin.de} \\}

\begin{document}
\maketitle

\begin{abstract}

Entity disambiguation (ED) is the task of linking mentions in text to corresponding entries in a knowledge base. Dual Encoders address this by embedding mentions and label candidates in a shared embedding space and applying a similarity metric to predict the correct label. In this work, we focus on evaluating key design decisions for Dual Encoder-based ED, such as its loss function, similarity metric, label verbalization format, and negative sampling strategy. We present the resulting model \approach, a document-level Dual Encoder model that includes contextual label verbalizations and efficient hard negative sampling. Additionally, we explore an iterative prediction variant that aims to improve the disambiguation of challenging data points. Comprehensive experiments on AIDA-Yago validate the effectiveness of our approach, offering insights into impactful design choices that result in a new State-of-the-Art system on the ZELDA benchmark.

\end{abstract}

\section{Introduction}

\begin{figure*}[h!] %[htpb]
    \centering
    \includegraphics[width=400px]{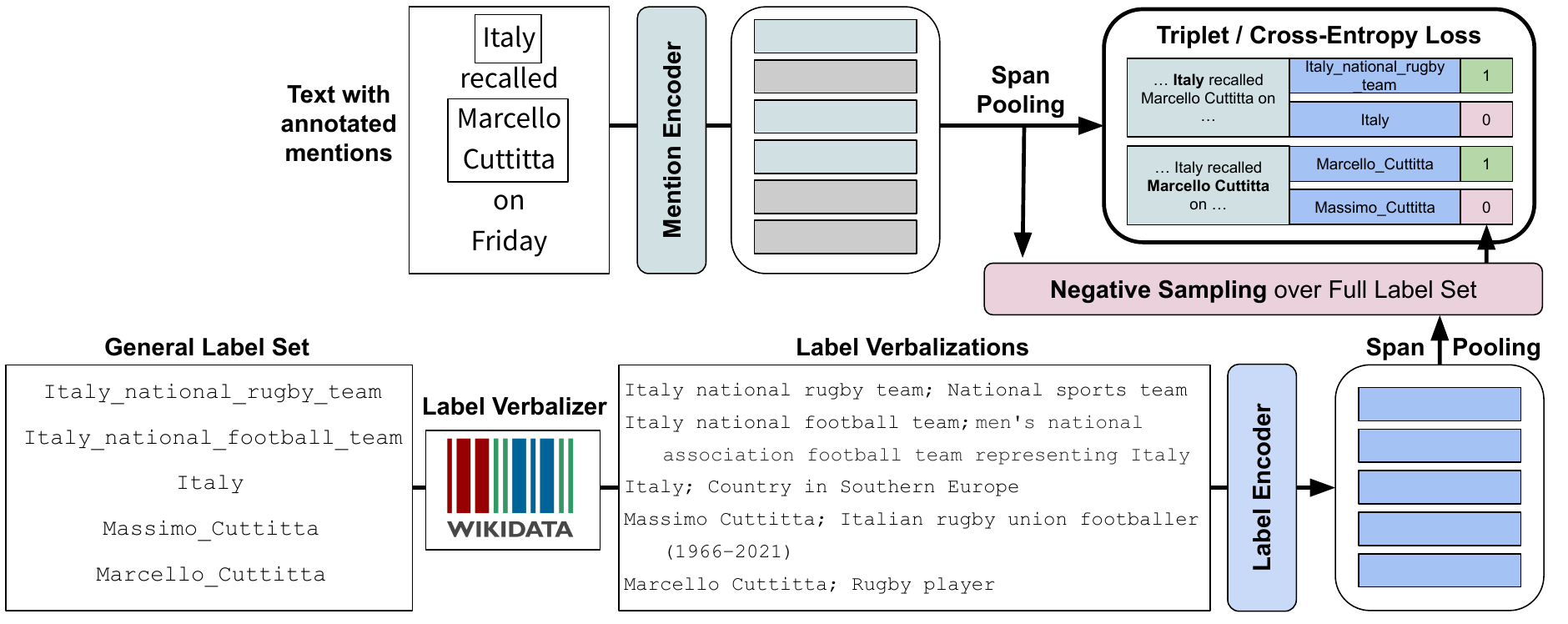}
    \caption{Overview of \approach~during training: The Mention Encoder produces an embedding for each entity mention in a given text (here "Italy" and "Marcello Cuttitta"). The Label Encoder similarly produces an embedding for each unique target in the General Label Set (spanning entities such as "Italy" and "Italy\_national\_football\_team"), by embedding their respective verbalizations. The purpose of training is to learn an embedding space in which mention embeddings lie close to the embeddings of the correct target verbalization. Training uses a Negative Sampling strategy which leverages embeddings to find hard negatives.}
    \label{fig:overview-approach}
\end{figure*}

Entity disambiguation (ED) is the task of resolving ambiguous mentions of named entities in text to their corresponding entries in a predefined knowledge base (KB), such as Wikipedia or Wikidata. The ability to correctly link mentions of entities (e.g., "Einstein" or "Princeton") to their respective KB entries is crucial for downstream tasks such as knowledge graph construction, question answering, and information retrieval. Formally, for a given set of entity mentions $\mathcal{M} = \{m_1, \ldots, m_T\}$ in corpus $\mathcal{D}$, entity disambiguation aims to link each mention $m_t$ to its corresponding gold entity $e_t$ from a set of possible entities $\mathcal{E} = \{e_1, \ldots, e_{|\mathcal{E}|}\}$.

Traditional ED systems often operate on lexical similarity, link popularity, hand-crafted features, and candidate lists \cite{ganea-hofmann-2017-deep, yamada-etal-2016-joint} or 
%which face issues like incompleteness, outdated entries, and limited applicability in low-resource domains or languages.
simple classification-based approaches \cite{broscheit-2019-investigating, févry2020empiricalevaluation} that involve fine-tuning a classification head on top of a pre-trained language model. More recent approaches are dense retrieval based, with the Dual Encoder \cite{gillick-etal-2019-learning, wu-etal-2020-scalable, procopio-etal-2023-entity, wang-etal-2024-entity} as one of the most popular architectures. It encodes mentions and KB entries into a shared vector space for similarity-based matching. 

However, despite its simplicity, the Dual Encoder architecture involves key design choices that can greatly influence its ability to properly disambiguate entities. For instance, key questions include how to best verbalize labels and how to model similarity. Furthermore, training is greatly affected by decisions on negative sampling, loss functions and efficient label embedding strategies.

\noindent 
\textbf{Contributions.}
Our work aims at evaluating those key design decisions for Dual Encoder models systematically. We further introduce \approach~as a resulting system, which integrates document-level processing, refined label verbalizations, hard negative sampling and efficient label embedding updates\footnote{We will release \approach~and our label verbalizations upon acceptance.}.
Additionally, we conduct exploratory experiments with an iterative prediction and verbalization strategy that leverages already predicted neighboring mentions. This approach aims to improve contextual understanding and address challenging ambiguous cases.

In more detail, our contributions are:
\begin{itemize}
  \item We present \approach, a dual encoder architecture for Entity Disambiguation that uses label verbalizations and efficient hard negative sampling, without relying on candidate lists.
  \item We conduct extensive experiments to evaluate our design choices on the AIDA-Yago benchmark and compare the resulting model to several other ED models on the much larger ZELDA benchmark.
  \item We introduce a variant of \approach~that predicts mentions iteratively, leveraging early disambiguations to assist in resolving challenging cases and show qualitative insights.

\end{itemize}

\section{The \approach~Architecture}
\label{approach}

This section outlines the Dual Encoder architecture and its key techniques and design decisions.

\subsection{Dual Encoder Basics}

Figure~\ref{fig:overview-approach} gives an overview of the Dual Encoder and its main components during training:

\begin{table*}[t]
\vspace{-3mm}
    \centering
    \small
    \begin{tabular}{p{0.10\textwidth} p{0.4\textwidth} p{0.4\textwidth}}
        \toprule
        Component & Albert\_Einstein & Wembley\_Stadium \\
        \midrule
        Title & Albert Einstein  & Wembley Stadium \\[2pt]
        Description & German-born theoretical physicist (1879–1955)  & football stadium in London, England \\[2pt]
        Categories & occupation: physicist, scientist  & instance of: multi-purpose sports venue; \\    
          &  & country: United Kingdom \\[2pt]
        Paragraph & Albert Einstein was a German-born theoretical physicist who is best known for developing the theory of relativity. [...]  & Wembley Stadium is an association football stadium in Wembley, London. It opened in 2007 on the site of the original Wembley Stadium
        %, which had stood from 1923 until 2003.
        [...] \\
        \bottomrule
    \end{tabular}
    \caption{Examples for different components for creating label verbalizations.}
    \label{verbalization-format-examples}
\vspace{-3mm}
\end{table*}

\paragraph{Mention Encoder.}
The Mention Encoder processes the textual context surrounding an entity mention. For instance, in a document or sentence containing the mention "Italy", the encoder considers the surrounding words to generate a contextually rich embedding for the mention. Our approach leverages the entire document as context during mention encoding, ensuring richer semantic information for each mention.

\paragraph{Label Encoder.}
The Label Encoder generates embeddings for all entities using metadata such as descriptions and KB relations for accurate representation. The label set may e.g. include "Italy" (the country), "Italy\_national\_rugby\_team" and "Italy\_national\_football\_team". The Label Verbalizer produces short descriptions for each, like verbalizing "Italy" as "Country in Southern Europe", which is then encoded into an entity embedding.

\paragraph{Similarity Computation.}
The embeddings from the Mention and Label Encoder are pooled to obtain span representations and then compared using a similarity metric (e.g., cosine similarity or Euclidean distance). The entity whose embedding is most similar to the mention embedding is selected as the predicted label. The purpose of training is thus to learn an embedding space in which mention embeddings lie close to the embedding of the correct target verbalization.

\paragraph{Sampling Negatives for Training.}
To improve disambiguation and training robustness, we incorporate negative sampling over the label pool, so the model learns to distinguish correct labels from hard negative candidates. For example, for the mention "Italy" with gold label "Italy\_national\_rugby\_team", a close but incorrect label like "Italy" (country) serves as negative label.

\paragraph{Inference.}
Inference involves embedding mentions and entities, comparing embeddings in the shared space and selecting the most similar match.

\begin{table}[t]
  %  \vspace{-3mm}
    \setlength{\tabcolsep}{0.1cm}
    \centering
        \begin{tabular}{lc}
            \toprule
            \textbf{Verbalization} & F1 \\
            \midrule
            Title & 63.68 ± 0.05 \\
            Title + Categories & 64.00 ± 0.13 \\
            Title + Description & 64.48 ± 0.10 \\
            Title + Description + Categories & \textbf{65.01 ± 0.08} \\
            Title + Paragraph (100) & 64.30 ± 0.02 \\
            Title + Paragraph (500) & 63.49 ± 0.50 \\
            \bottomrule
        \end{tabular}
        \caption{Comparing verbalizations formats.
        %Triplet loss, Euclidean, hard negatives, dynamic factor, mean pooling.
        %Default is soft cutoff after 50 characters, for Title+Par after 100 and 500 characters.
        }
        \label{tab:verbalization-format-experiments}
\vspace{-3mm}
\end{table}

% \subsection{Iterative \approach~ variant}
% \label{iterative-variant}

% We additionally introduce a variant of our base architecture that employs an \textit{iterative} prediction procedure during both training and inference. After predicting all mentions in a document, the N predictions with highest similarity score are selected. For those, we insert label verbalizations into the text, after their respective mentions. The resulting text version is then re-embedded, and the remaining mentions are re-predicted. This process repeats until all mentions are resolved\footnote{We set N to about one-third of mentions in the batch.}. This iterative approach aims to improve predictions, particularly for challenging ambiguous mentions, by incrementally enriching the context with entity descriptions. We refer to this variant as \textit{iterative}, while the base architecture is termed \textit{1-step}. We refer to the examples in Table \ref{iterative-examples-good} and \ref{iterative-examples-bad} (last columns) for an illustration of sentences with such label verbalization insertions.

% To adapt the training process to the iterative variant, we incorporate label verbalizations for a random subset of the occurring mentions also during training\footnote{Initially, we use gold labels (randomly corrupting 10\% for robustness) and later switch to confident predictions after a predefined number of update steps.}. The mentions that were selected for label verbalization insertion are excluded from loss calculation within the same batch, as their target label would be present in the text.

\subsection{Design Decisions for the Dual Encoder}
\label{design-choises}

The effectiveness of the Dual Encoder model relies on several key design choices, which we summarize here before providing detailed discussion and evaluation for each in Sect. \ref{ablation-experiments}.

\paragraph{Enriching representations.}
One of the most crucial aspects is creating high-quality entity representations. Expressive \textbf{label verbalizations}, like descriptions or structured KB data, enrich entity embeddings and help disambiguate context-sensitive mentions. These work alongside document-level mention representations to ensure rich semantic understanding. Both encoders require an effective \textbf{pooling strategy}, such as mean pooling, to obtain concise representations per mention or label.

\paragraph{Training Dynamics.}
Additionally, a suitable \textbf{similarity metric} is essential, as is a \textbf{loss function} to optimize the embedding space, i.e. pulling positive mention-entity pairs closer while pushing negatives away. Both require \textbf{negative samples}: Hard negatives -- incorrect entities similar to the mention -- help improve fine-grained differentiation, while in-batch negatives offer computational efficiency. For negative sampling, cached entity embeddings must be efficiently updated to reflect model changes, with the \textbf{update frequency} managed to balance accuracy and computation.

\section{Evaluating Design Choices}
\label{ablation-experiments}

We conduct several ablation experiments to assess the impact of the design choices sketched in Sect  \ref{design-choises}. Due to limited computational resources, these experiments were performed using the smaller AIDA-CoNLL-Yago \cite{hoffart-etal-2011-robust} as train set, while evaluation was carried out on the diverse, out-of-domain ZELDA test sets \cite{milich-akbik-2023-zelda}. The only exception is the ablation for label update frequency (Sect. \ref{updating-label-embeddings}), where we also trained on ZELDA. We report the mean F1 over ZELDA-test with standard deviation of three runs.
Unless otherwise specified, the default experimental setup uses Title + Description for verbalization, triplet loss, Euclidean distance, hard negative mining with a dynamic factor, and mean pooling, while varying the options of the respective design choice.

\subsection{Label Verbalization Formats}

\paragraph{Design Choices.}
We evaluate different verbalization formats and use Wikidata, similar to prior work \cite{procopio-etal-2023-entity, atzeni-etal-2023-polar}.
Table~\ref{verbalization-format-examples} illustrates the verbalizations we consider: 
(1) The entity's \textbf{Title}, (2) a short \textbf{Description}, or (3) more structured \textbf{Categories} using the \textit{instance\_of}, \textit{subset\_of}, \textit{country}, and \textit{occupation} relations from Wikidata. We further experiment with (4) using the first Wikipedia \textbf{Paragraph}. 
We test various component combinations, using a semicolon after the title and commas as separators. Verbalizations have a soft 50-character limit, splitting at the next punctuation. For paragraphs, we allow lengths of 100 or 500 characters.

\paragraph{Experimental Analysis.}
Table \ref{tab:verbalization-format-experiments} shows the results. Using only the title performs worst but remains competitive, suggesting that individual missing descriptions do not severely impact performance. Descriptions slightly outperform categories, but combining all three yields the best results. Descriptions add detail, while categories provide structure for better generalization. Since some entities have only descriptions (2.1\%) or categories (3.3\%), both are essential for full coverage. Wikipedia paragraphs perform worse, especially at the longer length of 500 characters.

\begin{table}[t]
  %  \vspace{-3mm}
    \centering
        \begin{tabular}{llc}
            \toprule
            Loss & \textbf{Pooling} & F1 \\
            \midrule
            Triplet & Mean & 64.48 ± 0.10 \\
            {} & First-last & \textbf{66.25 ± 0.40} \\
            \midrule
            Cross-Entropy & Mean & 65.84 ± 0.22 \\
            {} & First-last & \textbf{66.66 ± 0.09} \\
            \bottomrule
        \end{tabular}
        \caption{Comparing span pooling methods.
        %Hard negatives, dynamic negative factor, Euclidean, Title+Desc.
        }
        \label{tab:pooling-experiments}
\vspace{-3mm}
\end{table}

\subsection{Span Pooling Method}

\paragraph{Design Choices.} For both mention and label span representations, we evaluate two pooling methods: Taking the \textbf{mean} of the token embeddings within the mention span or label, or concatenating the embeddings of the \textbf{first and last} tokens of the mention span or label.
For label verbalizations, we use only title tokens, computing either their mean or concatenating the first and last token, while treating the rest (e.g. the description) as context, mirroring mention token processing.

\paragraph{Experimental Analysis.}
Table \ref{tab:pooling-experiments} shows that concatenating the embeddings of the first and last span token consistently performs better than averaging all span tokens. The first and last tokens often encapsulate critical boundary information of the span, which can be especially helpful for disambiguation.

\begin{table}[t]
    \vspace{-3mm}
    \centering
    \begin{tabular}{llc}
        \toprule
        \textbf{Loss} & \textbf{Similarity} & F1 \\
        \midrule
        Triplet & Cosine & 50.65 ± 0.20 \\
         & Dot Product & 64.43 ± 0.05 \\
         & Euclidean & 64.48 ± 0.10 \\
        \midrule
        Cross-Entropy & Cosine & 34.34 ± 0.25 \\
         & Dot Product & 64.52 ± 0.04 \\
         & Euclidean & \textbf{65.84 ± 0.22} \\
        \bottomrule
    \end{tabular}
    \caption{Comparing loss and similarity metrics.
    %Hard negatives, dynamic factor, mean pooling, Title+Desc.
    }
    \label{tab:losses-similarity-experiments}
\vspace{-3mm}
\end{table}

\subsection{Similarity Metric and Loss}
\label{losses-similarity}

\paragraph{Design Choices.}
The choice of similarity metric significantly impacts model effectiveness. We experiment with three options: \textbf{Cosine similarity}, which measures the angle between vectors and works well for normalized embeddings; \textbf{Euclidean Distance}, measuring the straight-line distance between vectors, used as a negative since our model is similarity-based; and \textbf{Dot Product}, which directly computes unnormalized similarity.

For optimizing the embedding space, we explore two loss functions: \textbf{Triplet Loss} pulls positive mention-label pairs closer while maintaining a margin from a given negative; \textbf{Cross-Entropy Loss} adapts the classification objective to entity disambiguation by aligning mention embeddings with correct labels and penalizing incorrect associations.

\paragraph{Experimental Analysis.}
Table \ref{tab:losses-similarity-experiments} shows that cross-entropy loss combined with Euclidean similarity achieves best performance. For triplet loss, both dot product and Euclidean distance yield similar results. Cosine distance performs worse.
%for both losses.

\subsection{Negative Sampling Methods}

\begin{table}[t]
    \vspace{-3mm}
    \centering
    \begin{tabular}{llc}
        \toprule
        Loss & \textbf{Negatives} & F1 \\
        \midrule
        Triplet & In-Batch, dyn & 54.39 ± 0.08 \\
        {} & Hard, 1 & 64.46 ± 0.06 \\
        {} & Hard, dyn & \textbf{64.48 ± 0.10} \\
        \midrule
        Cross-Entropy & In-Batch, dyn & 54.06 ± 0.14 \\
        {} & Hard, 1 & 65.78 ± 0.17 \\
        {} & Hard, dyn & \textbf{65.84 ± 0.22} \\
        \bottomrule
    \end{tabular}
    \caption{Comparing negative sampling methods.
    %Mean pooling, Euclidean, Title+Desc.
    }
    \label{tab:negative-sampling-experiments}
\vspace{-3mm}
\end{table}

\paragraph{Design Choices.}
Training with all possible entity candidates is computationally prohibitive, so we use negative sampling. For each mention, we either sample \textbf{in-batch negatives} (labels from other mentions in the batch) or \textbf{hard negatives} (incorrect entities most similar to the mention). The number of negatives per mention is another design choice.
As re-encoding all labels at each step is infeasible, negatives are retrieved from periodically refreshed cached embeddings, while gold and selected negatives are freshly embedded for loss calculation.

\paragraph{Experimental Analysis.}
The results in Table \ref{tab:negative-sampling-experiments} show that hard negative sampling significantly outperforms in-batch sampling. We compare using 1 negative label per positive sample and a \textit{dynamic} approach, where number of negatives is maximized based on GPU memory capacity for each batch, leading to marginal improvements.
%\footnote{Preliminary experiments with a higher number of negatives (e.g., 10) did not yield notable gains, therefore we excluded it from the ablations.}.

\subsection{Frequency of Label Embedding Updates}
\label{updating-label-embeddings}

\paragraph{Design Choices.}
Updating label embeddings is crucial for accurate representation, but re-encoding all labels every step is infeasible. To balance accuracy and efficiency, we cache embeddings and refresh them periodically, either \textbf{after each epoch} or at \textbf{more frequent} intervals. Additionally, labels actively used in a batch (positive or negative) are updated \textbf{on-the-fly}, keeping frequently used labels up to date without full re-encoding.

\paragraph{Experimental Analysis.}
We validate the effectiveness of frequently and dynamically updating cached label embeddings, rather than updating them only after each epoch (see Table \ref{tab:updating-label-embeddings-experiments}). Note that this ablation was conducted training on ZELDA, as the update frequency is more impactful for larger datasets where once every epoch is not enough. The results show that more frequent label embedding updates are crucial for performance.

\begin{table}[t]
  %  \vspace{-3mm}
    \centering
    \begin{tabular}{lc}
        \toprule
        \textbf{Label Embedding Updates} & F1 \\
        \midrule
        Once after Epoch        & 76.17 ± 0.04 \\
        Frequent + On-the-Fly   & \textbf{82.32 ± 0.10} \\
        \bottomrule
   \end{tabular}
   \caption{Comparing label embedding update frequency. Trained on \textbf{ZELDA}.
   %, triplet loss, Euclidean, hard negatives, dynamic factor, mean pooling, Title+Desc.
   }
   \label{tab:updating-label-embeddings-experiments}
\vspace{-3mm}
\end{table}

\begin{figure*}[h!] %[htpb]
\vspace{-3mm}
    \centering
    \includegraphics[width=400px]{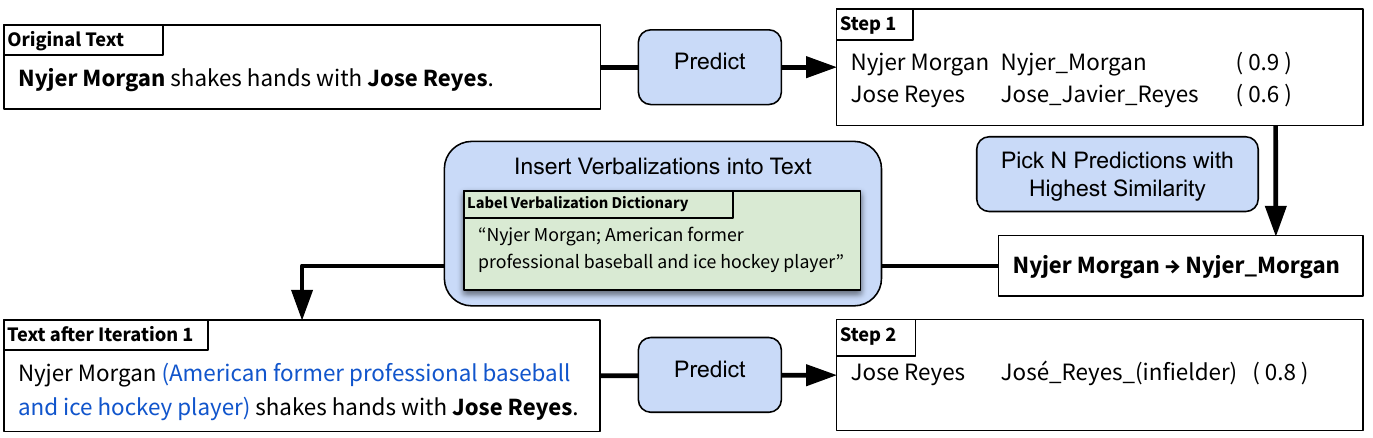}
    \caption{Iterative variant: All mentions are initially predicted, the top-N with highest similarity are selected for text insertion. The enriched text is then re-embedded, and the remaining mentions are re-predicted.}
    \label{fig:iterative-illustration}
\vspace{-3mm}
\end{figure*}

\subsection{Takeaways}
We summarize the key findings from the above ablations.
The best label verbalization format combines title, description, and categories, underlining the importance of semantic richness as well as coverage. For span pooling, concatenating the first and last token embeddings surpasses mean pooling. Among similarity metrics and loss functions, cross-entropy loss with Euclidean distance achieved the highest performance. Hard negatives consistently outperformed in-batch negatives. For large training data, frequent label embedding updates are crucial.
%These findings underscore the importance of expressive label information, robust pooling, and optimal label sampling and update strategies for generalizable disambiguation.

\section{Iterative Prediction}
\label{iterative-variant}

We experiment with an iterative prediction variant of our base architecture that aims to particularly help challenging, ambiguous cases.

\subsection{Enriching Context with Label Insertions}

In this approach, illustrated in Figure \ref{fig:iterative-illustration}, after predicting all mentions in a document, the N predictions with highest similarity scores are selected, and their label verbalizations are inserted into the text in brackets after their respective mentions. The modified text is re-embedded, and the remaining mentions are re-predicted. This process repeats until all mentions are resolved\footnote{N is set to one-third of the mentions in a batch. We only overwrite previous labels if the similarity score is higher.}. The goal is to incrementally enrich the context with entity descriptions, improving subsequent predictions. 

In the example from Figure \ref{fig:iterative-illustration}, the original text contains the mentions "Nyjer Morgan" and "Jose Reyes". In the first prediction step, the model identifies possible entities, assigning similarity scores (e.g., "Nyjer\_Morgan" with 0.9 and "Jose\_Javier\_Reyes" with 0.6). The most confident prediction, "Nyjer\_Morgan", is selected for text insertion, thus its verbalization is added to the text. In the next prediction step, this additional context helps the model correctly disambiguate "Jose Reyes" to "José\_Reyes\_(infielder)" with higher confidence (0.8) instead of the incorrect previous prediction, a film director. More examples of these insertions are shown in Tables \ref{iterative-examples-good} and \ref{iterative-examples-bad}.
%We refer to this variant as \textit{iterative prediction}.

\subsection{Modified Training}

Iterative insertion and prediction can, in theory, be directly applied during inference due to its natural language format. However, to reinforce its utility, we adapt the training process by incorporating label verbalizations for a random subset of mentions\footnote{Gold labels are used initially (10\% randomly corrupted), later in training we switch to confident predictions.}. Mentions selected for verbalization are excluded from the loss calculation of the current batch, as their target label is already present in the text.

\begin{table*}[ht]
    \vspace{-3mm}
    \centering
    \small
    \setlength{\tabcolsep}{0.07cm}

    \begin{tabular}{lccccccccccc}
        \toprule
        Method & \makecell{Train \\ data} & AIDA-B & TWEEKI & \makecell{REDDIT- \\ POSTS} & \makecell{REDDIT- \\ COMM} & \makecell{WNED-\\CWEB} & \makecell{WNED-\\WIKI} & \makecell{SLINKS-\\TAIL} & \makecell{SLINKS-\\SHAD.} & \makecell{SLINKS-\\TOP} & AVG \\
        %\midrule
        %\textit{CL-RECALL}\textsuperscript{1} & {} & 91.1 & 94.0 & 98.4 & 98.3 & 92.4 & 98.8 & 98.8 & 56.7 & 73.1 & 89.1 \\
        
        \midrule
        \multicolumn{10}{l}{\textit{Classification}} \\
        FEVRY$_{ALL}$ & {Z} & 79.2 & 71.8 & 88.5 & 84.1 & 68.0 & 84.3 & 63.8 & 43.4 & 53.1 & 70.7 \\
        FEVRY$_{CL}$ & {Z} & 79.5 & 76.9 & 89.0 & 86.5 & 70.3 & 84.5 & 87.6 & 31.9 & 47.7 & 72.7 \\
        LUKE$_{PRE}$ & {Z} & 79.3 & 73.8 & 76.1 & 69.9 & 66.8 & 68.4 & 97.7 & 20.4 & 50.8 & 67.0 \\
        LUKE$_{FT}$ & {Z} & 81.2 & 77.9 & 81.5 & 78.5 & 70.3 & 76.5 & 98.0 & 22.5 & 51.8 & 71.0 \\
        
        \midrule
        \multicolumn{10}{l}{\textit{Generative}} \\
        GENRE$_{ALL}$ & {Z} & 72.4 & 75.9 & 88.8 & 83.9 & 66.5 & 85.2 & 95.3 & 38.7 & 43.5 & 72.2 \\
        GENRE$_{CL}$ & {Z} & 78.6 & \underline{80.1} & \underline{92.8} & \underline{91.5} & \textbf{73.6} & 88.4 & \textbf{99.6} & 37.3 & 52.8 & 77.2 \\
        
        \midrule
        \multicolumn{10}{l}{\textit{Dense Retrieval}} \\
        \textsc{FusionED}  & {Z} & 80.1 & \textbf{81.4} & \textbf{93.9} & \textbf{92.3} & \textbf{73.6} & 89.0 & 98.3 & 41.5 & 57.9 & 78.7 \\
        
        \midrule
        \multicolumn{12}{l}{\textit{Proposed}} \\
    %    \multicolumn{12}{l}{\approach} \\

        \approach & {Z} & 82.6 & 78.9 & 89.2 & 86.2 & 69.8 & \textbf{91.4} & 98.6 & \underline{65.3} & \underline{67.0} & 81.0 \\  
        \hspace{0.3cm}\textit{+ iter. prediction} & {Z} & \underline{84.4} & 78.4 & 89.5 & 87.1 & 70.4 & \underline{91.2} & \underline{98.7} & \underline{65.3} & \textbf{67.2} & \underline{81.4} \\  
        \hspace{0.5cm}\textit{+ iter. training} & {Z} & \textbf{88.2} & 78.9 & 92.2 & 88.4 & \underline{71.5} & 90.8 & 98.2 & \textbf{66.3} & 65.9 & \textbf{82.3} \\  

        \bottomrule
    \end{tabular}
    \caption{Comparison between \approach~ and other SoTA models on the ZELDA benchmark.
    %, all trained on ZELDA train (Z).
    %\textit{Classification} and \textit{Generative} numbers are taken from \citet{milich-akbik-2023-zelda}, \textsc{FusionED} results from \citet{wang-etal-2024-entity}
    }

    \label{results-zelda}
\end{table*}

\section{Experiments on ZELDA benchmark}
\label{training-on-zelda}

We compare our resulting system \approach~to different baseline and competitive models.

\subsection{Experimental Setup}

\paragraph{Data.} The ZELDA benchmark \cite{milich-akbik-2023-zelda} addresses inconsistencies in previous ED setups by unifying training data and entity vocabularies. It includes 95,000 Wikipedia paragraphs (2.6M mentions, \textasciitilde822,000 unique entities) for training and nine test splits across diverse domains to evaluate model generalizability across varying text lengths and contexts. These are:

\textbf{AIDA-B} \cite{hoffart-etal-2011-robust}, the test split of the AIDA-Yago dataset, consisting of 231 news articles. \textbf{TWEEKI} \cite{harandizadeh-singh-2020-tweeki}, a collection of 500 tweets. \textbf{Reddit-POSTS} and \textbf{Reddit-COMMENTS}, a collection of posts and comments from the Reddit forum \cite{BOTZER2021102479}. \textbf{WNED-WIKI} and \textbf{WNED-CWEB}, a collection of Wikipedia articles vs. web pages \cite{Guo-Barbosa-Robust}. Finally, the \textbf{Shadowlinks} corpora, three datasets with different levels of difficulty in terms of \textit{overshadowing} \cite{provatorova-etal-2021-robustness}. \textbf{TOP} includes the easiest cases (the correct entity is the most frequent one), \textbf{SHADOW} the most difficult (the correct entity is overshadowed by a more frequent one) and \textbf{TAIL} includes generally rare though mostly not overshadowed entities.

In addition to the main experiments on ZELDA, we further evaluate on \textbf{MSNBC} \cite{cucerzan-2007-large}, \textbf{AQUAINT} \cite{mine-2008-learning} and \textbf{ACE2004} \cite{ratinov-etal-2011-local}, see Sect. \ref{ap-blink-chatel}.

\paragraph{General Label Set.}
Comparing ED approaches is challenging due to differences in pretraining, reliance on candidate lists and selected label set \cite{shavarani-sarkar-2023-spel, milich-akbik-2023-zelda, yamada-etal-2022-global, wang-etal-2024-entity, ong-etal-2024-unified}. For consistency, as a fixed label set we use all 821,402 unique ZELDA candidates and no mention specific candidate list.
Further training details and hyperparameters are in Appendix \ref{hyperparameter}.

\paragraph{Baselines and Competitive Models.}
We compare to the models reported by the ZELDA authors \cite{milich-akbik-2023-zelda}: Their reimplementation of the classification baseline \textbf{FEVRY} \cite{févry2020empiricalevaluation} with and without using the mention-candidate lists (FEVRY$_{CL}$ and FEVRY$_{ALL}$). A global ED model \textbf{LUKE} \cite{yamada-etal-2022-global}
in two variants: Pre-training in which entity embeddings are learned (LUKE$_{PRE}$), and an optional final epoch of fine-tuning with frozen entity embeddings (LUKE$_{FT}$). The generative model \textbf{GENRE} \cite{decao2021autoregressive}, which generates the target label title, restricting either to the full (GENRE$_{ALL}$) or to the mention's specific candidate pool (GENRE$_{CL}$). We add \textbf{\textsc{FusionED}} \cite{wang-etal-2024-entity}, a novel encoder-decoder architecture to fuse entity descriptions and candidate embeddings which is also trained on ZELDA.

We further compare our approach to \textbf{BLINK} \cite{wu-etal-2020-scalable} and \textbf{\textsc{ChatEL}} \cite{ding-etal-2024-chatel}. As differences in training data and label sets make direct comparison to our and the other models challenging, see Appendix \ref{ap-blink-chatel} for the results.

\subsection{Results}
\label{main-results}

We train \approach~on ZELDA train and evaluate on its test sets (Table \ref{results-zelda}). For MSNBC, ACE2004, AQUAINT, and comparison to BLINK and \textsc{ChatEL}, see Tables \ref{tab:results-blink} and \ref{tab:results-chatel-other-datasets} in \ref{ap-blink-chatel}.

\subsubsection{Baseline Comparison}

\noindent 
\textbf{Overall Performance.}
On average, \approach~achieves the highest performance over the ZELDA test sets, outperforming both classification and generative approaches. The highly competitive \textsc{FusionED} is surpassed on 5 of the 9 datasets.  

\begin{table*}[h!]
    \centering
    \small
    \setlength{\tabcolsep}{0.08cm}
    
    \begin{tabular}{p{0.09\textwidth} p{0.35\textwidth} p{0.53\textwidth}}
    \toprule
    {} & \textbf{\hspace{2cm}Step 1} & \textbf{\hspace{3cm}Step 2} \\
    
    \midrule
    Input & "Peggy Olson is awesome... one of the best characters on \#\textbf{madmen}" & "Peggy Olson \textcolor{blue}{(fictional character from Mad Men)} is awesome... one of the best characters on \#\textbf{madmen}" \\
    Prediction & Mad\_(TV\_series) \textcolor{red}{(\xmark)} &
    Mad\_Men \textcolor{teal}{(\cmark)} \\    

    \midrule
    Input & "Karlsruhe 3 (Reich 29th, Carl 44th, \textbf{Dundee} 69th) Freiburg 0. Halftime 2-0." & "Karlsruhe \textcolor{blue}{(German sport club)} 3 (Reich 29th, Carl 44th, \textbf{Dundee} 69th) Freiburg \textcolor{blue}{(German football club)} 0. Halftime 2-0." \\
    Prediction & Sean\_Dundee \textcolor{teal}{(\cmark)} &
    Dundee\_F.C. \textcolor{red}{(\xmark)} \\
        
    \bottomrule
    \end{tabular}
    \caption{Positive and negative example of label change after iterative prediction. See more in Tables \ref{iterative-examples-good} and \ref{iterative-examples-bad}.}
    \label{iterative_examples}
\vspace{-3mm}
\end{table*}
\noindent 
\textbf{Strength on Shadowlinks Corpora.}
\approach~achieves top results on SHADOW and TOP, where candidate list-based models struggle due to low recall (56.7\% on SHAD, 73.1\% on TOP). The candidate-independent models also face challenges with rare, ambiguous or overshadowed mentions. These results highlight \approach's ability to effectively leverage subtle contextual cues while also demonstrating its strength in not relying on mention-specific candidates.

\noindent 
\textbf{Performance on Long-Form Text.}
On AIDA-B and WNED-WIKI, \approach~tops all models and remains competitive on REDDIT-POSTS -- all datasets with long, continuous documents.

\noindent 
\textbf{Challenges with Short-Form Text.}
\approach~struggles on REDDIT-COMM and TWEEKI due to its reliance on long context, providing limited benefit for short social media posts. Still, it performs well, even surpassing other models on SHADOW, despite its single-sentence documents.

\noindent 
\textbf{Low Performance on WNED-CWEB.}
On WNED-CWEB, performance is lower than expected, comparable to classification models. We attribute this to web scraping artifacts and annotation inconsistencies\footnote{E.g., our model links "William Pitt" to a person, while gold is "University\_of\_Pittsburgh", or "taxpayers" to "Taxpayers" and "Rose" to "Derrick\_Rose," which seem more accurate than the gold labels "Tax" and the flower "Rose."}. While all models face these challenges, disjointed documents -- composed of unrelated sentences -- pose a greater challenge for document-based approaches like ours, which rely on coherent context for effective predictions.

\subsubsection{Evaluating the Iterative Variant}
\label{result-discussion-iterative}

For evaluating the iterative variant of \approach, we compare: (1) the base model without iteration, (2) the base model with iterative prediction (\textit{+ iter. prediction}), and (3) the full iterative variant incorporating both iterative training and prediction (\textit{+ iter. training}).
We observe some gains with iterative prediction on AIDA-B, REDDIT-COMM, and WNED-CWEB, while other datasets (TWEEKI, REDDIT-POSTS, WNED-WIKI, SLINKS) show no clear improvement. Adding insertions during training boosts performance. On average, and on five datasets, the iterative outperforms the base approach, though slight declines occur on WNED-WIKI, SLINKS-TAIL, and -TOP, and no difference on TWEEKI.
Shadowlinks datasets contain only single-mention documents, while TWEEKI and REDDIT-COMM, based on social media, include only few multi-mention documents. Since the iterative approach relies on neighboring mentions for label insertions, its effect is limited in these cases.

\paragraph{Qualitative Inspection.}
As the iterative variant showed only slight and inconsistent improvements over the base model, we qualitatively analyze iterative prediction steps. Table \ref{iterative_examples} provides examples for both positive and negative label change. As positive example, disambiguating "Peggy Olson" as a character from Mad Men aids in correctly labeling the series. Conversely, inserting two sports team labels leads to misinterpreting "Dundee" as the sports team "Dundee\_F.C." instead of the person "Sean\_Dundee", initially correctly predicted. Refer to \ref{appendix-qualitative-insights} for additional examples and discussion of both improved and erroneous predictions after label insertions, along with quantitative counts.

In summary, while the iterative approach helps with underspecified mentions and complex domains, its drawbacks -- susceptibility to linguistic patterns and error propagation -- outweigh its modest, inconsistent gains. Given the added training cost, we recommend the base \approach~architecture, already demonstrating strong performance.

\section{Related Work}
\label{related-work}

Entity Disambiguation (ED) models resolve ambiguous mentions in text to corresponding entities in a knowledge base (KB)\footnote{While Entity \textit{Disambiguation} (ED) assumes mentions already identified, Entity \textit{Linking} (EL) also detects spans. We focus on ED, though it is often integrated into end-to-end EL or Relation Extraction systems \cite{shavarani-sarkar-2023-spel, bouziani-etal-2024-rexel, orlando-etal-2024-relik}.}. Recent advances use transformers for better contextual representations and zero-shot abilities. %This section reviews key approaches and their relation to our work.

\paragraph{Candidates.} ED is often split into candidate retrieval and ranking \cite{procopio-etal-2023-entity, wu-etal-2020-scalable, yamada-etal-2022-global, yang-etal-2019-learning}. Many methods use precompiled candidate lists (alias tables) to map mentions to small sets of potential entities, reducing the search space \cite{procopio-etal-2023-entity, yamada-etal-2022-global, wang-etal-2024-entity, yang-etal-2019-learning}. Common lists are e.g. by \citet{ganea-hofmann-2017-deep} or \citet{le-titov-2018-improving}.

\paragraph{Simple Classifiers.}
Early classification-based ED approaches, like \citet{broscheit-2019-investigating} and \cite{févry2020empiricalevaluation}, use a softmax classification head on a pretrained transformer. While effective, they are limited by the need for a fixed, small label set and sufficient training data for each label, making adaptation to new domains or labels challenging.

\paragraph{Dense Retrieval and Ranking.}
To provide more flexibility in label sets, Dense Retrieval has gained popularity, embedding mentions and entity candidates into a shared embedding space. Many models further enrich label representations by incorporating expressive label descriptions \cite{procopio-etal-2023-entity, wu-etal-2020-scalable, provatorova2022nice}.

\textbf{Dual Encoders} (or Bi-Encoders) \cite{gillick-etal-2019-learning, humeau2020polyencoders} embed mentions and entities into a shared space using separate encoders, predicting labels via similarity. While often used for retrieval \cite{orlando-etal-2024-relik}, they can also make final predictions. \textbf{Cross Encoders} rank a small set of pre-retrieved candidates
%by methods like Dual Encoders, candidate lists, or string matching
by jointly encoding mentions and candidates, capturing deeper interactions for improved ranking.
Both are often combined: Efficient candidate retrieval with a Dual Encoder, then ranking via Cross Encoder, like in the BLINK model \cite{wu-etal-2020-scalable}. For training the Bi-Encoder, they mainly use in-batch negatives, optionally adding hard ones.
%BLINK uses precomputed top-10 predictions per mention, while we sample negatives dynamically per batch.
While BLINK's Bi-Encoder focuses on candidate retrieval and prioritizes recall, our Dual Encoder directly handles label prediction, omitting the expensive Cross Encoder step.
\citet{wang-etal-2024-entity} introduce a novel encoder-decoder architecture \textsc{FusionED}, which employs an encoder-decoder architecture to learn interactions between the text and each candidate entity.
%Then the decoder fuses candidate representations for selecting the label.
While using entity descriptions and hard negative sampling is similar to our work, its decoder-based approach introduces more complexity.

Many approaches further enhance dense retrieval models by incorporating structural knowledge from knowledge graphs such as entity types or relations \cite{ayoola-etal-2022-refined, atzeni-etal-2023-polar, bouziani-etal-2024-rexel}, similarly to \citet{provatorova2022nice} and \citet{tedeschi-etal-2021-named-entity}, who use entity type information for filtering candidates.

\paragraph{Global Prediction.}
Most ED approaches, including our base model, treat mentions individually (local ED). Our iterative variant instead inserts dynamic label verbalizations from neighboring mentions, aligning it with global ED approaches \cite{yamada-etal-2022-global, oba-etal-2022-entity, xiao-etal-2023-coherent, ganea-hofmann-2017-deep, le-titov-2018-improving, yang-etal-2018-collective, provatorova2022nice, yang-etal-2019-learning}.
LUKE \cite{yamada-etal-2022-global} proposes a global ED model that resolves mentions within a document sequentially, using previously resolved entities as additional input tokens. Unlike our verbalization insertions, LUKE appends entity titles and relies on candidate lists. %LUKE focuses on optimizing sequential predictions, while our approach aims to disambiguate all mentions in fewer collective iterations.

\paragraph{Alternative Methods.} Generative models like GENRE \cite{decao2021autoregressive} frame ED as sequence-to-sequence to generate label titles from an entity list. Other methods use knowledge graphs \cite{li-2022-improving-entity-linking}, structured prediction \cite{shavarani-sarkar-2023-spel}, or span extraction \cite{barba-etal-2022-extend}. Recently, few-shot ED with LLM prompting gained more attention \cite{liu-etal-2024-onenet, zhou-etal-2024-gendecider, xu-etal-2023-read, ding-etal-2024-chatel}.

\section{Conclusion}

We systematically evaluated key design choices for Dual Encoder-based entity disambiguation, focusing on loss functions, similarity metrics, label verbalization formats, negative sampling and efficient label embedding updates. Our experiments on the AIDA and ZELDA benchmarks provide valuable insights into the impact of these decisions for the effectiveness of Dual Encoder models for ED.

Based on these investigations, we introduced \approach, a system that integrates document-level processing, contextual label verbalizations, efficient hard negative sampling, and cashed label embeddings and achieves state-of-the-art performance on the ZELDA benchmark. By eliminating the reliance on candidate lists, \approach~ offers a scalable and flexible solution.

While our iterative prediction strategy on average improved performance, gains were inconsistent and qualitative analysis revealed some unwanted negative effects. However, with positive cases mostly prevailing, there is potential for further pursuing this variant.

\pagebreak
\section*{Limitations}
\label{sec:limitations}

While our approach demonstrates significant advancements in entity disambiguation, several limitations remain that offer avenues for future improvement:

\paragraph{Limited Evaluation of Training Configurations.} Due to time and resource constraints, we could not run extensive training experiments on the ZELDA dataset across multiple seeds or hyperparameter configurations and had to rely on the ablations conducted on AIDA-Yago. This limits the robustness of our reported results on this benchmark. It is possible, that the difference in size and diversity of ZELDA train, would favor different settings for certain design choices.
Furthermore, there are several hyperparameters that we did not evaluate systematically, such as the margin parameter for triplet loss, which may significantly impact its effectiveness depending on the chosen similarity metric or more in-depth analysis of the effect of the number of negative samples per datapoint, length of label verbalizations or, concerning the iterative prediction variant, insertion format.

\paragraph{Dependency on Descriptions for Labels.} Our approach relies heavily on the availability of detailed descriptions for most labels in the entity set. This dependence may restrict its applicability to domains or datasets lacking such. On the other hand, our approach \textit{only} relies on descriptions and does not require e.g. candidate lists which also are challenging to collect. Furthermore, the title-only setting still showed reasonable results, suggesting that individual sparse cases pose minimal concern.

\paragraph{Training Complexity of the Iterative Model.} The iterative variant of our approach for enriching entity disambiguation with contextual verbalizations shows potential, but a) requires significantly more time and computational resources to train and also for inference, and b) given the inconclusive results concerning its added value, further investigation is needed to fully understand its impact, refine its methodology and determine its usefulness.

\paragraph{Focus on English Datasets.} Our experiments were conducted exclusively on English ED datasets, leaving the generalizability of our method to multilingual scenarios unexplored. Unfortunately, while very valuable, this was beyond the scope of this work. Thaid said, applying dual encoder systems to other languages could pose challenges including the quality of multilingual resources (e.g., Wikidata), linguistic differences which might affect span encoding method (e.g., first-last pooling), and the scarcity of large-scale annotated datasets. Future work addressing these challenges will help ensure broader applicability.

\section*{Ethical Considerations}
\label{sec:ethics-statement}

While our approach to entity disambiguation has few direct ethical concerns, some considerations arise due to inherent limitations in language models and the reliance on external data. Human-written label descriptions can contain errors, misinformation, and biases. Potentially, rare entities, being less documented, might be more likely to be misclassified, perpetuating disparities in representation. Additionally, the language model may perpetuate biases, such as gender stereotyping in professions or favoring frequent entities \cite{chen-etal-2021-evaluating, provatorova-etal-2021-robustness}.

\section*{Acknowledgements}

We thank all reviewers for their valuable comments. Susanna Rücker and Alan Akbik are supported by the Deutsche Forschungsgemeinschaft (DFG, German Research Foundation) under Emmy Noether grant "Eidetic Representations of Natural Language" (project number 448414230). Further, Alan Akbik is supported by the Deutsche Forschungsgemeinschaft (DFG, German Research Foundation) under Germany’s Excellence Strategy "Science of Intelligence" (EXC 2002/1, project number 390523135).

We acknowledge the use of AI generative tools that assisted with language, presentation, and formulation of this paper. These tools were utilized to refine the phrasing, improve clarity, and ensure the linguistic quality of the document. The ideas, content, and research presented in this paper are entirely our own, and these tools were not involved in generating the scientific concepts or findings discussed. All concepts, methodologies, and interpretations are original and derived from our work.

% Entries for the entire Anthology, followed by custom entries

\bibliography{anthology,custom}

\begin{thebibliography}{41}
\expandafter\ifx\csname natexlab\endcsname\relax\def\natexlab#1{#1}\fi

\bibitem[{Atzeni et~al.(2023)Atzeni, Plekhanov, Dreyer, Kassner, Merello,
  Martin, and Cancedda}]{atzeni-etal-2023-polar}
Mattia Atzeni, Mikhail Plekhanov, Frederic Dreyer, Nora Kassner, Simone
  Merello, Louis Martin, and Nicola Cancedda. 2023.
\newblock \href {https://doi.org/10.18653/v1/2023.emnlp-main.566} {Polar ducks
  and where to find them: Enhancing entity linking with duck typing and polar
  box embeddings}.
\newblock In \emph{Proceedings of the 2023 Conference on Empirical Methods in
  Natural Language Processing}, pages 9129--9146, Singapore. Association for
  Computational Linguistics.

\bibitem[{Ayoola et~al.(2022)Ayoola, Tyagi, Fisher, Christodoulopoulos, and
  Pierleoni}]{ayoola-etal-2022-refined}
Tom Ayoola, Shubhi Tyagi, Joseph Fisher, Christos Christodoulopoulos, and
  Andrea Pierleoni. 2022.
\newblock \href {https://doi.org/10.18653/v1/2022.naacl-industry.24}
  {{R}e{F}in{ED}: An efficient zero-shot-capable approach to end-to-end entity
  linking}.
\newblock In \emph{Proceedings of the 2022 Conference of the North American
  Chapter of the Association for Computational Linguistics: Human Language
  Technologies: Industry Track}, pages 209--220, Hybrid: Seattle, Washington +
  Online. Association for Computational Linguistics.

\bibitem[{Barba et~al.(2022)Barba, Procopio, and
  Navigli}]{barba-etal-2022-extend}
Edoardo Barba, Luigi Procopio, and Roberto Navigli. 2022.
\newblock \href {https://doi.org/10.18653/v1/2022.acl-long.177} {{E}xt{E}n{D}:
  Extractive entity disambiguation}.
\newblock In \emph{Proceedings of the 60th Annual Meeting of the Association
  for Computational Linguistics (Volume 1: Long Papers)}, pages 2478--2488,
  Dublin, Ireland. Association for Computational Linguistics.

\bibitem[{Botzer et~al.(2021)Botzer, Ding, and Weninger}]{BOTZER2021102479}
Nicholas Botzer, Yifan Ding, and Tim Weninger. 2021.
\newblock \href {https://doi.org/https://doi.org/10.1016/j.ipm.2020.102479}
  {Reddit entity linking dataset}.
\newblock \emph{Information Processing \& Management}, 58(3):102479.

\bibitem[{Bouziani et~al.(2024)Bouziani, Tyagi, Fisher, Lehmann, and
  Pierleoni}]{bouziani-etal-2024-rexel}
Nacime Bouziani, Shubhi Tyagi, Joseph Fisher, Jens Lehmann, and Andrea
  Pierleoni. 2024.
\newblock \href {https://doi.org/10.18653/v1/2024.naacl-industry.11} {{REXEL}:
  An end-to-end model for document-level relation extraction and entity
  linking}.
\newblock In \emph{Proceedings of the 2024 Conference of the North American
  Chapter of the Association for Computational Linguistics: Human Language
  Technologies (Volume 6: Industry Track)}, pages 119--130, Mexico City,
  Mexico. Association for Computational Linguistics.

\bibitem[{Broscheit(2019)}]{broscheit-2019-investigating}
Samuel Broscheit. 2019.
\newblock \href {https://doi.org/10.18653/v1/K19-1063} {Investigating entity
  knowledge in {BERT} with simple neural end-to-end entity linking}.
\newblock In \emph{Proceedings of the 23rd Conference on Computational Natural
  Language Learning (CoNLL)}, pages 677--685, Hong Kong, China. Association for
  Computational Linguistics.

\bibitem[{Chen et~al.(2021)Chen, Gudipati, Longpre, Ling, and
  Singh}]{chen-etal-2021-evaluating}
Anthony Chen, Pallavi Gudipati, Shayne Longpre, Xiao Ling, and Sameer Singh.
  2021.
\newblock \href {https://doi.org/10.18653/v1/2021.acl-long.345} {Evaluating
  entity disambiguation and the role of popularity in retrieval-based {NLP}}.
\newblock In \emph{Proceedings of the 59th Annual Meeting of the Association
  for Computational Linguistics and the 11th International Joint Conference on
  Natural Language Processing (Volume 1: Long Papers)}, pages 4472--4485,
  Online. Association for Computational Linguistics.

\bibitem[{Cucerzan(2007)}]{cucerzan-2007-large}
Silviu Cucerzan. 2007.
\newblock \href {https://aclanthology.org/D07-1074} {Large-scale named entity
  disambiguation based on {W}ikipedia data}.
\newblock In \emph{Proceedings of the 2007 Joint Conference on Empirical
  Methods in Natural Language Processing and Computational Natural Language
  Learning ({EMNLP}-{C}o{NLL})}, pages 708--716, Prague, Czech Republic.
  Association for Computational Linguistics.

\bibitem[{{De Cao} et~al.(2021){De Cao}, Izacard, Riedel, and
  Petroni}]{decao2021autoregressive}
Nicola {De Cao}, Gautier Izacard, Sebastian Riedel, and Fabio Petroni. 2021.
\newblock \href {https://openreview.net/forum?id=5k8F6UU39V} {Autoregressive
  entity retrieval}.
\newblock In \emph{9th International Conference on Learning Representations,
  {ICLR} 2021, Virtual Event, Austria, May 3-7, 2021}. OpenReview.net.

\bibitem[{Devlin et~al.(2019)Devlin, Chang, Lee, and
  Toutanova}]{devlin-etal-2019-bert}
Jacob Devlin, Ming-Wei Chang, Kenton Lee, and Kristina Toutanova. 2019.
\newblock \href {https://doi.org/10.18653/v1/N19-1423} {{BERT}: Pre-training of
  deep bidirectional transformers for language understanding}.
\newblock In \emph{Proceedings of the 2019 Conference of the North {A}merican
  Chapter of the Association for Computational Linguistics: Human Language
  Technologies, Volume 1 (Long and Short Papers)}, pages 4171--4186,
  Minneapolis, Minnesota. Association for Computational Linguistics.

\bibitem[{Ding et~al.(2024)Ding, Zeng, and Weninger}]{ding-etal-2024-chatel}
Yifan Ding, Qingkai Zeng, and Tim Weninger. 2024.
\newblock \href {https://aclanthology.org/2024.lrec-main.275/} {{C}hat{EL}:
  Entity linking with chatbots}.
\newblock In \emph{Proceedings of the 2024 Joint International Conference on
  Computational Linguistics, Language Resources and Evaluation (LREC-COLING
  2024)}, pages 3086--3097, Torino, Italia. ELRA and ICCL.

\bibitem[{Févry et~al.(2020)Févry, FitzGerald, Soares, and
  Kwiatkowski}]{févry2020empiricalevaluation}
Thibault Févry, Nicholas FitzGerald, Livio~Baldini Soares, and Tom
  Kwiatkowski. 2020.
\newblock \href {http://arxiv.org/abs/2005.14253} {Empirical evaluation of
  pretraining strategies for supervised entity linking}.

\bibitem[{Ganea and Hofmann(2017)}]{ganea-hofmann-2017-deep}
Octavian-Eugen Ganea and Thomas Hofmann. 2017.
\newblock \href {https://doi.org/10.18653/v1/D17-1277} {Deep joint entity
  disambiguation with local neural attention}.
\newblock In \emph{Proceedings of the 2017 Conference on Empirical Methods in
  Natural Language Processing}, pages 2619--2629, Copenhagen, Denmark.
  Association for Computational Linguistics.

\bibitem[{Gillick et~al.(2019)Gillick, Kulkarni, Lansing, Presta, Baldridge,
  Ie, and Garcia-Olano}]{gillick-etal-2019-learning}
Daniel Gillick, Sayali Kulkarni, Larry Lansing, Alessandro Presta, Jason
  Baldridge, Eugene Ie, and Diego Garcia-Olano. 2019.
\newblock \href {https://doi.org/10.18653/v1/K19-1049} {Learning dense
  representations for entity retrieval}.
\newblock In \emph{Proceedings of the 23rd Conference on Computational Natural
  Language Learning (CoNLL)}, pages 528--537, Hong Kong, China. Association for
  Computational Linguistics.

\bibitem[{Guo and Barbosa(2017)}]{Guo-Barbosa-Robust}
Zhaochen Guo and Denilson Barbosa. 2017.
\newblock \href {https://doi.org/10.3233/SW-170273} {Robust named entity
  disambiguation with random walks}.
\newblock \emph{Semantic Web}, 9:1--21.

\bibitem[{Harandizadeh and Singh(2020)}]{harandizadeh-singh-2020-tweeki}
Bahareh Harandizadeh and Sameer Singh. 2020.
\newblock \href {https://doi.org/10.18653/v1/2020.wnut-1.29} {Tweeki: Linking
  named entities on {T}witter to a knowledge graph}.
\newblock In \emph{Proceedings of the Sixth Workshop on Noisy User-generated
  Text (W-NUT 2020)}, pages 222--231, Online. Association for Computational
  Linguistics.

\bibitem[{Hoffart et~al.(2011)Hoffart, Yosef, Bordino, F{\"u}rstenau, Pinkal,
  Spaniol, Taneva, Thater, and Weikum}]{hoffart-etal-2011-robust}
Johannes Hoffart, Mohamed~Amir Yosef, Ilaria Bordino, Hagen F{\"u}rstenau,
  Manfred Pinkal, Marc Spaniol, Bilyana Taneva, Stefan Thater, and Gerhard
  Weikum. 2011.
\newblock \href {https://aclanthology.org/D11-1072} {Robust disambiguation of
  named entities in text}.
\newblock In \emph{Proceedings of the 2011 Conference on Empirical Methods in
  Natural Language Processing}, pages 782--792, Edinburgh, Scotland, UK.
  Association for Computational Linguistics.

\bibitem[{Humeau et~al.(2020)Humeau, Shuster, Lachaux, and
  Weston}]{humeau2020polyencoders}
Samuel Humeau, Kurt Shuster, Marie-Anne Lachaux, and Jason Weston. 2020.
\newblock \href {http://arxiv.org/abs/1905.01969} {Poly-encoders: Transformer
  architectures and pre-training strategies for fast and accurate
  multi-sentence scoring}.

\bibitem[{Le and Titov(2018)}]{le-titov-2018-improving}
Phong Le and Ivan Titov. 2018.
\newblock \href {https://doi.org/10.18653/v1/P18-1148} {Improving entity
  linking by modeling latent relations between mentions}.
\newblock In \emph{Proceedings of the 56th Annual Meeting of the Association
  for Computational Linguistics (Volume 1: Long Papers)}, pages 1595--1604,
  Melbourne, Australia. Association for Computational Linguistics.

\bibitem[{Li et~al.(2022)Li, Li, Li, Li, Liu, Liu, and
  Dong}]{li-2022-improving-entity-linking}
Qijia Li, Feng Li, Shuchao Li, Xiaoyu Li, Kang Liu, Qing Liu, and Pengcheng
  Dong. 2022.
\newblock \href {https://doi.org/10.3390/app12052702} {Improving entity linking
  by introducing knowledge graph structure information}.
\newblock \emph{Applied Sciences}, 12(5).

\bibitem[{Liu et~al.(2024)Liu, Liu, Zhang, Wang, Liu, and
  Chen}]{liu-etal-2024-onenet}
Xukai Liu, Ye~Liu, Kai Zhang, Kehang Wang, Qi~Liu, and Enhong Chen. 2024.
\newblock \href {https://doi.org/10.18653/v1/2024.emnlp-main.756} {{O}ne{N}et:
  A fine-tuning free framework for few-shot entity linking via large language
  model prompting}.
\newblock In \emph{Proceedings of the 2024 Conference on Empirical Methods in
  Natural Language Processing}, pages 13634--13651, Miami, Florida, USA.
  Association for Computational Linguistics.

\bibitem[{Milich and Akbik(2023)}]{milich-akbik-2023-zelda}
Marcel Milich and Alan Akbik. 2023.
\newblock \href {https://aclanthology.org/2023.eacl-main.151} {{ZELDA}: A
  comprehensive benchmark for supervised entity disambiguation}.
\newblock In \emph{Proceedings of the 17th Conference of the European Chapter
  of the Association for Computational Linguistics}, pages 2061--2072,
  Dubrovnik, Croatia. Association for Computational Linguistics.

\bibitem[{Milne and Witten(2008)}]{mine-2008-learning}
David Milne and Ian~H. Witten. 2008.
\newblock \href {https://doi.org/10.1145/1458082.1458150} {Learning to link
  with wikipedia}.
\newblock In \emph{Proceedings of the 17th ACM Conference on Information and
  Knowledge Management}, CIKM '08, page 509–518, New York, NY, USA.
  Association for Computing Machinery.

\bibitem[{Oba et~al.(2022)Oba, Yamada, Yoshinaga, and
  Toyoda}]{oba-etal-2022-entity}
Daisuke Oba, Ikuya Yamada, Naoki Yoshinaga, and Masashi Toyoda. 2022.
\newblock \href {https://aclanthology.org/2022.findings-emnlp.472} {Entity
  embedding completion for wide-coverage entity disambiguation}.
\newblock In \emph{Findings of the Association for Computational Linguistics:
  EMNLP 2022}, pages 6333--6344, Abu Dhabi, United Arab Emirates. Association
  for Computational Linguistics.

\bibitem[{Ong et~al.(2024)Ong, Shavarani, and Sarkar}]{ong-etal-2024-unified}
Nicolas Ong, Hassan Shavarani, and Anoop Sarkar. 2024.
\newblock \href {https://doi.org/10.18653/v1/2024.naacl-short.11} {Unified
  examination of entity linking in absence of candidate sets}.
\newblock In \emph{Proceedings of the 2024 Conference of the North American
  Chapter of the Association for Computational Linguistics: Human Language
  Technologies (Volume 2: Short Papers)}, pages 113--123, Mexico City, Mexico.
  Association for Computational Linguistics.

\bibitem[{Orlando et~al.(2024)Orlando, Huguet~Cabot, Barba, and
  Navigli}]{orlando-etal-2024-relik}
Riccardo Orlando, Pere-Llu{\'\i}s Huguet~Cabot, Edoardo Barba, and Roberto
  Navigli. 2024.
\newblock \href {https://doi.org/10.18653/v1/2024.findings-acl.839}
  {{R}e{L}i{K}: Retrieve and {L}in{K}, fast and accurate entity linking and
  relation extraction on an academic budget}.
\newblock In \emph{Findings of the Association for Computational Linguistics:
  ACL 2024}, pages 14114--14132, Bangkok, Thailand. Association for
  Computational Linguistics.

\bibitem[{Procopio et~al.(2023)Procopio, Conia, Barba, and
  Navigli}]{procopio-etal-2023-entity}
Luigi Procopio, Simone Conia, Edoardo Barba, and Roberto Navigli. 2023.
\newblock \href {https://aclanthology.org/2023.eacl-main.93} {Entity
  disambiguation with entity definitions}.
\newblock In \emph{Proceedings of the 17th Conference of the European Chapter
  of the Association for Computational Linguistics}, pages 1297--1303,
  Dubrovnik, Croatia. Association for Computational Linguistics.

\bibitem[{Provatorova et~al.(2021)Provatorova, Bhargav, Vakulenko, and
  Kanoulas}]{provatorova-etal-2021-robustness}
Vera Provatorova, Samarth Bhargav, Svitlana Vakulenko, and Evangelos Kanoulas.
  2021.
\newblock \href {https://doi.org/10.18653/v1/2021.emnlp-main.820} {Robustness
  evaluation of entity disambiguation using prior probes: the case of entity
  overshadowing}.
\newblock In \emph{Proceedings of the 2021 Conference on Empirical Methods in
  Natural Language Processing}, pages 10501--10510, Online and Punta Cana,
  Dominican Republic. Association for Computational Linguistics.

\bibitem[{Provatorova et~al.(2022)Provatorova, Tedeschi, Vakulenko, Navigli,
  and Kanoulas}]{provatorova2022nice}
Vera Provatorova, Simone Tedeschi, Svitlana Vakulenko, Roberto Navigli, and
  Evangelos Kanoulas. 2022.
\newblock \href {http://arxiv.org/abs/2210.06164} {Focusing on context is nice:
  Improving overshadowed entity disambiguation}.

\bibitem[{Ratinov et~al.(2011)Ratinov, Roth, Downey, and
  Anderson}]{ratinov-etal-2011-local}
Lev Ratinov, Dan Roth, Doug Downey, and Mike Anderson. 2011.
\newblock \href {https://aclanthology.org/P11-1138} {Local and global
  algorithms for disambiguation to {W}ikipedia}.
\newblock In \emph{Proceedings of the 49th Annual Meeting of the Association
  for Computational Linguistics: Human Language Technologies}, pages
  1375--1384, Portland, Oregon, USA. Association for Computational Linguistics.

\bibitem[{Shavarani and Sarkar(2023)}]{shavarani-sarkar-2023-spel}
Hassan Shavarani and Anoop Sarkar. 2023.
\newblock \href {https://doi.org/10.18653/v1/2023.emnlp-main.686} {{S}p{EL}:
  Structured prediction for entity linking}.
\newblock In \emph{Proceedings of the 2023 Conference on Empirical Methods in
  Natural Language Processing}, pages 11123--11137, Singapore. Association for
  Computational Linguistics.

\bibitem[{Tedeschi et~al.(2021)Tedeschi, Conia, Cecconi, and
  Navigli}]{tedeschi-etal-2021-named-entity}
Simone Tedeschi, Simone Conia, Francesco Cecconi, and Roberto Navigli. 2021.
\newblock \href {https://doi.org/10.18653/v1/2021.findings-emnlp.220} {{N}amed
  {E}ntity {R}ecognition for {E}ntity {L}inking: {W}hat works and what{'}s
  next}.
\newblock In \emph{Findings of the Association for Computational Linguistics:
  EMNLP 2021}, pages 2584--2596, Punta Cana, Dominican Republic. Association
  for Computational Linguistics.

\bibitem[{Wang et~al.(2024)Wang, Mousavi, Attia, Pradeep, Potdar, Rush, Minhas,
  and Li}]{wang-etal-2024-entity}
Junxiong Wang, Ali Mousavi, Omar Attia, Ronak Pradeep, Saloni Potdar, Alexander
  Rush, Umar~Farooq Minhas, and Yunyao Li. 2024.
\newblock \href {https://doi.org/10.18653/v1/2024.naacl-long.363} {Entity
  disambiguation via fusion entity decoding}.
\newblock In \emph{Proceedings of the 2024 Conference of the North American
  Chapter of the Association for Computational Linguistics: Human Language
  Technologies (Volume 1: Long Papers)}, pages 6524--6536, Mexico City, Mexico.
  Association for Computational Linguistics.

\bibitem[{Wu et~al.(2020)Wu, Petroni, Josifoski, Riedel, and
  Zettlemoyer}]{wu-etal-2020-scalable}
Ledell Wu, Fabio Petroni, Martin Josifoski, Sebastian Riedel, and Luke
  Zettlemoyer. 2020.
\newblock \href {https://doi.org/10.18653/v1/2020.emnlp-main.519} {Scalable
  zero-shot entity linking with dense entity retrieval}.
\newblock In \emph{Proceedings of the 2020 Conference on Empirical Methods in
  Natural Language Processing (EMNLP)}, pages 6397--6407, Online. Association
  for Computational Linguistics.

\bibitem[{Xiao et~al.(2023)Xiao, Shou, Zhang, Wu, Gong, and
  Jiang}]{xiao-etal-2023-coherent}
Zilin Xiao, Linjun Shou, Xingyao Zhang, Jie Wu, Ming Gong, and Daxin Jiang.
  2023.
\newblock \href {https://doi.org/10.18653/v1/2023.findings-emnlp.502} {Coherent
  entity disambiguation via modeling topic and categorical dependency}.
\newblock In \emph{Findings of the Association for Computational Linguistics:
  EMNLP 2023}, pages 7480--7492, Singapore. Association for Computational
  Linguistics.

\bibitem[{Xu et~al.(2023)Xu, Chen, Hu, and Zhang}]{xu-etal-2023-read}
Zhenran Xu, Yulin Chen, Baotian Hu, and Min Zhang. 2023.
\newblock \href {https://doi.org/10.18653/v1/2023.findings-emnlp.912} {A
  read-and-select framework for zero-shot entity linking}.
\newblock In \emph{Findings of the Association for Computational Linguistics:
  EMNLP 2023}, pages 13657--13666, Singapore. Association for Computational
  Linguistics.

\bibitem[{Yamada et~al.(2016)Yamada, Shindo, Takeda, and
  Takefuji}]{yamada-etal-2016-joint}
Ikuya Yamada, Hiroyuki Shindo, Hideaki Takeda, and Yoshiyasu Takefuji. 2016.
\newblock \href {https://doi.org/10.18653/v1/K16-1025} {Joint learning of the
  embedding of words and entities for named entity disambiguation}.
\newblock In \emph{Proceedings of the 20th {SIGNLL} Conference on Computational
  Natural Language Learning}, pages 250--259, Berlin, Germany. Association for
  Computational Linguistics.

\bibitem[{Yamada et~al.(2022)Yamada, Washio, Shindo, and
  Matsumoto}]{yamada-etal-2022-global}
Ikuya Yamada, Koki Washio, Hiroyuki Shindo, and Yuji Matsumoto. 2022.
\newblock \href {https://doi.org/10.18653/v1/2022.naacl-main.238} {Global
  entity disambiguation with {BERT}}.
\newblock In \emph{Proceedings of the 2022 Conference of the North American
  Chapter of the Association for Computational Linguistics: Human Language
  Technologies}, pages 3264--3271, Seattle, United States. Association for
  Computational Linguistics.

\bibitem[{Yang et~al.(2019)Yang, Gu, Lin, Tang, Zhuang, Wu, Chen, Hu, and
  Ren}]{yang-etal-2019-learning}
Xiyuan Yang, Xiaotao Gu, Sheng Lin, Siliang Tang, Yueting Zhuang, Fei Wu,
  Zhigang Chen, Guoping Hu, and Xiang Ren. 2019.
\newblock \href {https://doi.org/10.18653/v1/D19-1026} {Learning dynamic
  context augmentation for global entity linking}.
\newblock In \emph{Proceedings of the 2019 Conference on Empirical Methods in
  Natural Language Processing and the 9th International Joint Conference on
  Natural Language Processing (EMNLP-IJCNLP)}, pages 271--281, Hong Kong,
  China. Association for Computational Linguistics.

\bibitem[{Yang et~al.(2018)Yang, Irsoy, and Rahman}]{yang-etal-2018-collective}
Yi~Yang, Ozan Irsoy, and Kazi~Shefaet Rahman. 2018.
\newblock \href {https://doi.org/10.18653/v1/N18-1071} {Collective entity
  disambiguation with structured gradient tree boosting}.
\newblock In \emph{Proceedings of the 2018 Conference of the North {A}merican
  Chapter of the Association for Computational Linguistics: Human Language
  Technologies, Volume 1 (Long Papers)}, pages 777--786, New Orleans,
  Louisiana. Association for Computational Linguistics.

\bibitem[{Zhou et~al.(2024)Zhou, Li, Wang, Qiao, and
  Li}]{zhou-etal-2024-gendecider}
Kang Zhou, Yuepei Li, Qing Wang, Qiao Qiao, and Qi~Li. 2024.
\newblock \href {https://doi.org/10.18653/v1/2024.naacl-short.22}
  {{G}en{D}ecider: Integrating {``}none of the candidates{''} judgments in
  zero-shot entity linking re-ranking}.
\newblock In \emph{Proceedings of the 2024 Conference of the North American
  Chapter of the Association for Computational Linguistics: Human Language
  Technologies (Volume 2: Short Papers)}, pages 239--245, Mexico City, Mexico.
  Association for Computational Linguistics.

\end{thebibliography}
\bibliographystyle{acl_natbib}

\appendix

\section{Appendix}

\subsection{Hyperparameters for Training}
\label{hyperparameter}

We use bert-base-uncased \cite{devlin-etal-2019-bert} as the backbone for both the label and mention encoders, each comprising 110 million parameters\footnote{We chose BERT-base due to its balance between computational efficiency and strong performance. Using more powerful models like BERT-large or RoBERTa could potentially improve results by providing richer contextual embeddings, but these models come with significantly higher computational and memory requirements. Given the scale of training (\textasciitilde300 model runs for our ablations) and the document-level context used in this work, BERT-base was a practical choice. However, we argue that using larger models might not necessarily result in significant gains. LLMs excel on tasks requiring extensive reasoning, but for tasks like ED, where representations rely heavily on structured context and fine-tuning, the improvements might be marginal. Also, in Dual Encoder setups, the simplicity of embedding space operations (like similarity comparisons) may not fully leverage the additional representational power of larger models.}.

Models are trained with a batch size of 32 documents per update step. To handle CUDA memory limitations, especially with long documents in the ZELDA benchmark, we employ minibatching based on maximum chunk length and number of mentions per chunk: When mention count or text length exceeds GPU capacity, documents are split into smaller chunks, each containing up to 100 mentions and 2,800 characters, while maintaining as much contextual continuity as possible. Verbalizations are truncated after a soft threshold of 50 (100 or 500 for the paragraph setting) characters, using heuristics to avoid splitting words or phrases. Label embeddings are computed with a batch size of 128. Training employs the AdamW optimizer with a learning rate of 5e-6. For the triplet loss, margin values are set to 0.5 for cosine similarity and 3.0 for euclidean and dot product similarity.

Label embeddings are fully updated at the start of each epoch. For the large ZELDA training set, additional full updates occur after every 160,000 spans to prevent outdated negative samples, as well as the on-the-fly updates (see Sect. \ref{updating-label-embeddings}).

In training the iterative variant, we initially insert a random subset of gold labels per batch (partly corrupted). After 30,000 spans, we switch to inserting real predictions instead, aligning the training setup with the procedure during inference. 

\subsection{Qualitative Insights of Iterative Prediction}
\label{appendix-qualitative-insights}

\begin{table*}[ht]
    \centering
    \small
    
    \begin{tabular}{p{0.07\textwidth} p{0.35\textwidth} p{0.5\textwidth}}
    \toprule
    {} & \textbf{\hspace{3cm}Step 1} & \textbf{\hspace{3cm}Step 2} \\
    \midrule
    Input excerpt & "A few more days until I can watch \textbf{Penn} state get bent over and rammed by Alabama lol" & "A few more days until I can watch \textbf{Penn} state get bent over and rammed by Alabama \textcolor{blue}{(University of Alabama Football Team)} lol" \\
    Pred. & Pennsylvania\_State\_University \textcolor{red}{(\xmark)} &
    Penn\_State\_Nittany\_Lions\_football \textcolor{teal}{(\cmark)} \\
    Verbal. & "Pennsylvania State University; public university in Pennsylvania" & "Penn State Nittany Lions football; football team of Penn State University" \\
    
    \midrule
    Input excerpt & "Nyjer Morgan makes \textbf{Jose Reyes} seem tolerable." & "Nyjer Morgan \textcolor{blue}{(American former professional baseball and ice hockey player)} makes \textbf{Jose Reyes }seem tolerable." \\
    Pred. & Jose\_Javier\_Reyes \textcolor{red}{(\xmark)} &
    José\_Reyes\_(infielder) \textcolor{teal}{(\cmark)} \\    
    Verbal. & "José Javier Reyes, Filipino writer and director for film and television" & "José Reyes (infielder); Dominican baseball player, MLB All-Star" \\
    
    \midrule
    Input excerpt & "Relations between Clarke, \textbf{Major} good - spokesman. LONDON 1996-12-06 Relations between Chancellor of the Exchequer Kenneth Clarke and Prime Minister John Major are good despite media reports” & "Relations between Clarke, \textbf{Major} good - spokesman. LONDON \textcolor{blue}{(capital and largest city of England and the United Kingdom}) 1996-12-06 Relations between Chancellor of the Exchequer Kenneth Clarke \textcolor{blue}{(British politician (born 1940))} and Prime Minister John Major \textcolor{blue}{(former prime minister of the United Kingdom (born 1943))} are good despite media reports” \\
    Pred. & Major \textcolor{red}{(\xmark)} &
    John\_Major \textcolor{teal}{(\cmark)} \\    
    Verbal. & "Major" & "John Major; former prime minister of the United Kingdom (born 1943)" \\

    \midrule
    Input excerpt & "Peggy Olson is awesome... one of the best characters on \#\textbf{madmen}" & "Peggy Olson \textcolor{blue}{(fictional character from Mad Men)} is awesome... one of the best characters on \#\textbf{madmen}" \\
    Pred. & Mad\_(TV\_series) \textcolor{red}{(\xmark)} &
    Mad\_Men \textcolor{teal}{(\cmark)} \\    
    Verbal. & "Mad; American adult animated sketch comedy television series" & "Mad Men; American period drama television series" \\

    \midrule
    Input excerpt & "the director of \textbf{law and order}, his name is Dick Wolf" & "the director of \textbf{law and order}, his name is Dick Wolf \textcolor{blue}{(American television producer (born 1946))}" \\
    Pred. & Law\_and\_order \textcolor{red}{(\xmark)} & Law\_\&\_Order \textcolor{teal}{(\cmark)} \\
    Verbal. & "Law and Order; Wikimedia disambiguation page" & "Law \& Order; American police procedural and legal drama television series" \\
    
    \bottomrule
    \end{tabular}
    \caption{Insights to the iterative variant: Examples for successful disambiguation after label insertions.}
    \label{iterative-examples-good}
\end{table*}

\begin{table*}[ht]
    \centering
    \small
    
    \begin{tabular}{p{0.07\textwidth} p{0.35\textwidth} p{0.5\textwidth}}
    \toprule
    {} & \textbf{\hspace{3cm}Step 1} & \textbf{\hspace{3cm}Step 2} \\
    
    \midrule
    Input excerpt & "Karlsruhe 3 (Reich 29th, Carl 44th, \textbf{Dundee} 69th) Freiburg 0. Halftime 2-0. Attendance 33,000" & "Karlsruhe \textcolor{blue}{(German sport club)} 3 (Reich 29th, Carl 44th, \textbf{Dundee} 69th) Freiburg \textcolor{blue}{(German football club)} 0. Halftime 2-0. Attendance 33,000" \\
    Pred. & Sean\_Dundee \textcolor{teal}{(\cmark)} &
    Dundee\_F.C. \textcolor{red}{(\xmark)} \\
    Verbal. & "Sean Dundee; South African–German footballer" & "Dundee F.C.; association football club in Dundee, Scotland" \\
    
    \midrule
    Input excerpt & "West Indies tour manager Clive Lloyd has apologised for Lara's behaviour on Tuesday . He (\textbf{Lara}) had told Australia coach Geoff Marsh that ..." & "West Indies \textcolor{blue}{(multinational cricket team)} tour manager Clive Lloyd has apologised for Lara's behaviour on Tuesday . He (\textbf{Lara}) had told Australia \textcolor{blue}{(national sports team)} coach Geoff Marsh that ..." \\
    Pred. & Brian\_Lara \textcolor{teal}{(\cmark)} & Australia\_national\_cricket\_team \textcolor{red}{(\xmark)} \\
    Verbal. & "Brian Lara; West Indian cricketer" & "Australia national cricket team; national sports team" \\
    
    \midrule
    Input excerpt & "We offer the following types of posters; Classic Film posters, movie posters, French movie posters, \textbf{Italian} movie posters, cinema posters, affiche de cinema, bogart posters [...] Vintage Movie posters about the greats of their time like Humphrey Bogart, Marylin Monroe, Audrey Hepburn, Brigitte Bardot, Marlene Dietrich, James Dean, Greta Garbo [...]" & "We offer the following types of posters; Classic Film posters, movie posters, French movie posters, \textbf{Italian} movie posters, cinema posters, affiche de cinema, bogart posters [...] Vintage Movie posters about the greats of their time like Humphrey Bogart \textcolor{blue}{(American actor (1899–1957))}, Marylin Monroe, Audrey Hepburn \textcolor{blue}{(British actress (1929–1993))}, Brigitte Bardot, Marlene Dietrich, James Dean \textcolor{blue}{(American actor (1931–1955))}, Greta Garbo \textcolor{blue}{(Swedish-American actress (1905–1990))} [...]" \\
    Pred. & Italy \textcolor{teal}{(\cmark)} & Cinema\_of\_Italy \textcolor{red}{(\xmark)} \\
    Verbal. & "Italy; country in Southern Europe" & "Cinema of Italy; aspect of history" \\

    \midrule
    Input excerpt & "Alright. ESPN not in \textbf{HD} has sound. HD doesn't. Boo." & "Alright. ESPN \textcolor{blue}{(American television and radio sports network)} not in \textbf{HD} has sound. HD doesn't. Boo." \\
    Pred. & High-definition\_television \textcolor{teal}{(\cmark)} & HD\_Radio \textcolor{red}{(\xmark)} \\
    Verbal. & "High-definition television; TV resolution standard" & "HD Radio; digital radio technology" \\
    
    \bottomrule
    \end{tabular}
    \caption{Insights to the iterative variant: Examples for detrimental label insertions resulting in worse prediction.}
    \label{iterative-examples-bad}
\end{table*}

In addition to the short discussion in Sect. \ref{result-discussion-iterative}, we give some more qualitative insights into the model predictions of the iterative \approach~variant.

\paragraph{Examples for Label Change.}
Table \ref{iterative-examples-good} highlights instances where label insertions improved predictions. In the first example, "Penn" was initially mislinked to Pennsylvania State University but corrected to the Penn State football team after inserting "University of Alabama Football Team" from a neighboring mention. Similarly, the context of baseball clarified "Jose Reyes" as a Dominican baseball player rather than a film and television director.

However, the iterative approach did not deliver consistent performance improvements. Table \ref{iterative-examples-bad} illustrates cases where insertions misled predictions. E.g., when multiple insertions referenced sports teams ("German sport club", "German football club"), the model over-relied on these, favoring sports-related predictions even when the mention referred to a person ("Dundee\_F.C" vs. "Sean\_Dundee"). Similarly, when many insertions include the terms "actress" and "actor", this leads to the mention "Italian" being linked to "Cinema\_of\_Italy" instead of the more general "Italy" (country). While both solutions could be argued correct in this instance, in other cases we see real errors due to too much reliance or "mimicking" of the insertions. While we attempted to address this with a modified iterative training setup that includes a mix of mentions with and without disambiguated neighbouring mentions, the results remained mixed, as the positive and negative effects balance out.

\paragraph{Quantitative Counts.}
In Table \ref{iterative-prediction-changes} we present a quantitative analysis of the iterative variant’s impact on prediction quality, comparing the initial and final prediction state. We categorize instances into four key types: "Correct", where both initial and final predictions match the gold label, "Incorrect", where both differ from the gold label, "Incorrect > Correct", where the initial incorrect prediction changes to a correct one after iterative adjustments, and "Correct > Incorrect", where the initial correct prediction later changes to an incorrect one.

We observe that as expected, no changes occur for the Shadowlinks datasets, as they consist solely of single-mention documents, and only a few changes are seen for social-media-based datasets with mostly short documents with few mentions. Although positive changes are generally more common than negative ones, there are still a significant number of negative changes. Overall, the rate of change is low, with most labels remaining unchanged.

These analyses highlight the mixed results and modest benefits of the iterative approach, particularly considering its increased costs and complexity. However, it also suggests that there is potential in using enriched contextual clues, though the negative effects and surprisingly small gains would need to be addressed before confidently proposing this variant as the primary model. Still, we view the successful cases as evidence of the approach’s potential and the failures as valuable lessons for developing future global or iterative ED models.

\begin{table*}[ht]
    \centering
    \small
    \setlength{\tabcolsep}{0.08cm}

    \begin{tabular}{lccccccccc}
        \toprule
        Method  & AIDA-B & TWEEKI & \makecell{REDDIT- \\ POSTS} & \makecell{REDDIT- \\ COMM} & \makecell{WNED-\\CWEB} & \makecell{WNED-\\WIKI} & \makecell{SLINKS-\\TAIL} & \makecell{SLINKS-\\SHAD.} & \makecell{SLINKS-\\TOP} \\
        \midrule
        correct           &  3843 &  657 &      642 &     561 &   7771 &  6054 &     884 &       599 &    595 \\
        incorrect > correct &    97 &   10 &        7 &       2 &    171 &    70 &       0 &         0 &      0 \\
        correct > incorrect &    63 &   12 &        2 &       4 &    156 &    79 &       0 &         0 &      0 \\
        incorrect         &   482 &  177 &       52 &      70 &   3042 &   562 &      15 &       305 &    309 \\
        \#Mentions       &  4485 &  856 &      703 &     637 &  11140 &  6765 &     899 &       904 &    904 \\
        \midrule
        Accuracy Step 1    &     87.1 &    \textbf{78.2} &        91.6 &       \textbf{88.7} &      71.2 &     \textbf{90.7} &       98.3 &         66.3 &      65.8 \\
        Accuracy Last Step &     \textbf{87.9} &    77.9 &        \textbf{92.3} &       88.4 &      \textbf{71.3} &     90.5 &       98.3 &         66.3 &      65.8 \\
        \bottomrule
    \end{tabular}
    \caption{Insights to the iterative variant's performance over iteration steps. \textit{Correct}: initial and final prediction align with gold label, \textit{incorrect}: initial and final prediction are different from gold label, \textit{incorrect > correct} and \textit{correct > incorrect}: prediction changes from initially incorrect to correct and vice versa.}
\label{iterative-prediction-changes}
\end{table*}

\subsection{Comparison to BLINK and \textsc{ChatEL}}
\label{ap-blink-chatel}

We add two competitive models to our comparison, which we exclude from our main results (Table \ref{results-zelda}) due to differences in training and evaluation settings, making direct comparison challenging.

\paragraph{Comparison to BLINK.}

\begin{table*}[ht]
    \centering
    \small
    \setlength{\tabcolsep}{0.08cm}

    \begin{tabular}{lccccccccccc}
        \toprule
        Method & \makecell{Train \\ data} & AIDA-B & TWEEKI & \makecell{REDDIT- \\ POSTS} & \makecell{REDDIT- \\ COMM} & \makecell{WNED-\\CWEB} & \makecell{WNED-\\WIKI} & \makecell{SLINKS-\\TAIL} & \makecell{SLINKS-\\SHAD.} & \makecell{SLINKS-\\TOP} & AVG \\
        \midrule
        
        BLINK-OG$_{bi}$\textsuperscript{1} & {B} & 80.6 & 77.3 & 90.8 & 87.8 & 68.2 & 79.8 & 97.9 & 50.1 & 57.3 & 76.6 \\
        BLINK-OG$_{cross}$\textsuperscript{1} & {B} & \underline{84.2} & \textbf{82.4} & \textbf{92.8} & \textbf{91.2} & \textbf{77.3} & 82.3 & \textbf{99.2} & 64.8 & \textbf{74.2} & \textbf{83.2} \\
        BLINK-Z$_{bi}$\textsuperscript{2} & {Z} & 65.5 & 72.1 & 83.1 & 79.1 & 58.1 & 73.1 & 96.3 & 41.8 & 42.6 & 68.0 \\
        
        \midrule
        \textit{\approach} & {Z} & 82.6 & \underline{78.9} & 89.2 & 86.2 & 69.8 & \textbf{91.4} & \underline{98.6} & \underline{65.3} & \underline{67.0} & 81.0 \\  

        \hspace{0.5cm}\textit{+ iter. training} & {Z} & \textbf{88.2} & \underline{78.9} & \underline{92.2} & \underline{88.4} & \underline{71.5} & \underline{90.8} & 98.2 & \textbf{66.3} & 65.9 & \underline{82.3} \\

    \bottomrule
    \end{tabular}
    \caption{Comparison between \approach~and BLINK on the ZELDA benchmark.
    \textsuperscript{1}BLINK-OG results come from evaluating the original BLINK on ZELDA test sets \cite{wu-etal-2020-scalable}. However, they are not fully comparable since BLINK-OG was trained on significantly more data (B).
    \textsuperscript{2}BLINK-Z: We trained a BLINK Bi-Encoder on ZELDA, using our general label set with the BLINK verbalizations.
    }
    \label{tab:results-blink}

\end{table*}

BLINK \cite{wu-etal-2020-scalable} combines a Bi-Encoder with a subsequent Cross Encoder. The Bi-Encoder retrieves top 100 candidates using dot product similarity, which are then ranked by the the Cross Encoder. As the Bi-Encoder is very similar to our Dual Encode (the biggest difference being the verbalizations and negative sampling) and BLINK is currently widely used for ED, comparison remains interesting. 

As the BLINK model was trained on different data and with a different label set\footnote{The BLINK dataset consist of 9M datapoints with a label dictionary of size 5.9M, created from a 2019 Wikipedia dump \cite{wu-etal-2020-scalable}. ZELDA only includes 2.6M datapoints and 822K entities \cite{milich-akbik-2023-zelda}.}, direct comparison to our approach is challenging. See Table \ref{tab:results-blink} for the following comparison setups, for all of which we used their code base\footnote{\url{https://github.com/facebookresearch/BLINK}}:
We first evaluate their final model on the ZELDA test sets, both the Bi-Encoder (BLINK-OG$_{bi}$) and the Cross Encoder (BLINK-OG$_{cross}$).
For better comparison, we also train a BLINK Bi-Encoder on ZELDA (BLINK-Z$_{bi}$), where we plug in our general label set (ca. 800K entities) but keep the BLINK verbalizations\footnote{Note that \textasciitilde3\% of the labels in our label set do not appear in the BLINK label set. We exclude those from the label set for this experiment as well as exclude the affected datapoints from evaluation.}.

Unsurprisingly the original BLINK-OG$_{cross}$ -- benefiting from more data and the additional cross-encoder step -- beats \approach~on most (not all) datasets and on average, while our iterative variant comes close. Interestingly, our model (also the base variant) significantly outperforms the original BLINK Bi-Encoder. This is particularly impressive given the similar architecture of BLINK's Bi-Encoder and its advantage in training data\footnote{However we acknowledged that their bigger label set makes accurate disambiguation also more challenging.}.

For the BLINK Bi-Encoder trained on ZELDA, the results lag behind our Dual Encoder as well as the original BLINK Bi-Encoder\footnote{We did not have the resources to perform an extensive hyperparameter search for BLINK. Furthermore, based on our observations and reports from an open issue (\url{github.com/facebookresearch/BLINK/issues/31}), the code version in the repository appears to rely solely on in-batch negatives, omitting the hard negatives described in \citet{wu-etal-2020-scalable}. This discrepancy suggests that the original BLINK model may have benefited from a more robust negative sampling strategy, which we could not reproduce.}. Next to the smaller train data, this performance drop is likely due to differences in BLINK's and our training setup like our dynamic hard negatives, large context, constant label embedding updates and more concise label verbalizations, all of which likely contribute to its superior performance. 

\paragraph{BLINK's vs. our Label Verbalizations.}

For direct comparison of the effect of the applied label verbalizations, we trained two BLINK Bi-Encoders on AIDA with our general label set: Once with our verbalizations (Title+Description), and once with the BLINK verbalizations (first Wikipedia paragraph).
The results in Table \ref{blink-on-aida} highlight the effectiveness of concise descriptions in otherwise exact settings, moreover when only having access to a small train set like AIDA.

\begin{table*}
    \centering
    \small
    \setlength{\tabcolsep}{0.06cm}
    \begin{tabular}{lccccccccccc}
        \toprule
        Method & \makecell{Train \\ data} & AIDA-B & TWEEKI & \makecell{REDDIT- \\ POSTS} & \makecell{REDDIT- \\ COMM} & \makecell{WNED-\\CWEB} & \makecell{WNED-\\WIKI} & \makecell{SLINKS-\\TAIL} & \makecell{SLINKS-\\SHAD.} & \makecell{SLINKS-\\TOP} & AVG \\
        \midrule
        \multicolumn{12}{l}{BLINK$_{bi}$ with} \\
        \hspace{0.3cm}
        BLINK verbalizations & A & 59.4 & 47.9 & 50.9 & 53.7 & 44.2 & 50.5 & 76.3 & \textbf{40.3} & 38.5 & 51.3 \\
        \hspace{0.3cm}
        our verbalizations & A & \textbf{61.8}  & \textbf{57.7} & \textbf{77.4 } &\textbf{ 73.2} & \textbf{49.5} & \textbf{51.8 }& \textbf{80.1} & 23.3 & \textbf{25.3} & \textbf{55.6} \\
        \midrule
    \end{tabular}
    \caption{Training a BLINK Bi-Encoder on AIDA, using our label set with BLINK's vs. our verbalizations.}
    \label{blink-on-aida}
\end{table*}

\paragraph{Comparison to \textsc{ChatEL}.}

\begin{table*}[ht]
    \centering
    \small
    \setlength{\tabcolsep}{0.08cm}

    \begin{tabular}{lcccccccc}
        \toprule
        Method & \makecell{Train \\ data} & AIDA & CWEB & WIKI & MSNBC & ACE2004 & AQUAINT & AVG \\
        \midrule
        \multicolumn{9}{l}{\textit{LLM Prompting}} \\
        \textsc{ChatEL} & (B+CL) & 82.1 & 71.1 & 77.1 & \underline{86.6} & \underline{88.4} & 79.1 & 80.7 \\
        \midrule
        \multicolumn{9}{l}{\textit{Dense Retrieval}} \\
        %\rowcolor{Gray}
        BLINK-OG$_{bi}$ & B & 80.6 & 68.2 & 79.8 & 83.5 & 84.3 & \underline{87.2} & 80.6 \\
        %\rowcolor{Gray}
        BLINK-OG$_{cross}$ & B & \underline{84.2} & \textbf{77.3} & 82.3 & \textbf{97.1} & \textbf{98.4} & \textbf{98.7} & \textbf{89.7} \\
        BLINK-Z$_{bi}$ & Z & 65.5 & 58.1 & 73.1 & 67.4 & 74.0 & 79.8 & 69.7 \\
        \midrule
        \multicolumn{9}{l}{\approach} \\
        \hspace{0.3cm}\textit{ZELDA labels} & Z & 82.6 & 69.8 & \textbf{91.4} & 74.7 & 75.9 & 73.0 & 77.9 \\
        \hspace{0.3cm}\textit{+ additional labels} & Z & 82.6 & 69.8 & \textbf{91.4} & 80.3 & 82.5 & 80.5 & 81.2 \\
        \hspace{0.5cm}\textit{+ iter. training} & Z & \textbf{88.2} & \underline{71.5} & \underline{90.8} & 80.8 & 85.6 & 84.2 & \underline{83.5} \\
        \bottomrule
    \end{tabular}
    \caption{\approach~ performance compared to \textsc{ChatEL} and BLINK. We include the datasets for which both models report numbers, including MSNBC, ACE2004, and AQUAINT. \textit{(B+CL)}: \textsc{ChatEL}, as a prompting method, does not use any specific train data, however for candidate generation they use both BLINK as well as candidate lists based on frequency statistics from hyperlinks \cite{ganea-hofmann-2017-deep}. \textit{+ additional labels}: About 15\% of labels from MSNBC, ACE2004 and AQUAINT are not included in our ZELDA-based label set. To enable fair evaluation, we add those to the label set for inference.}
    \label{tab:results-chatel-other-datasets}
\end{table*}

We also include a representative of another system type, \textsc{ChatEL} \cite{ding-etal-2024-chatel}, a current method that leverages LLM prompting for ED. In this approach, a small set of entity candidates is provided to a LLM. It is first tasked to enhance the mention context by generating auxiliary information from the document and its own knowledge and, in a second step, it is asked to select the correct entity through a multiple-choice formatted prompt.

\begin{table*}[!]
    \centering
    \small
    \setlength{\tabcolsep}{0.08cm}
    \begin{tabular}{lcccccccccc}
        \toprule
        Method  & AIDA-B & TWEEKI & \makecell{REDDIT- \\ POSTS} & \makecell{REDDIT- \\ COMM} & \makecell{WNED-\\CWEB} & \makecell{WNED-\\WIKI} & \makecell{SLINKS-\\TAIL} & \makecell{SLINKS-\\SHAD.} & \makecell{SLINKS-\\TOP} & AVG \\
        
        \midrule
        
        \approach & 94.5 & 91.0 & 97.7 & 95.9 & 83.1 & \textbf{96.3} & 99.6 & 97.7 & \textbf{93.9} & 94.4 \\
        
        \hspace{0.3cm}\textit{+ iter. training} & \textbf{95.7} & \textbf{91.1} & \textbf{97.9} & \textbf{97.3} & \textbf{84.0} & 95.6 & \textbf{99.7} & \textbf{98.1} & 92.6 & \textbf{94.7} \\
        
        \bottomrule
    \end{tabular}
    \caption{Results on ZELDA, restricting to the respective target dataset's label set for inference.}
    \label{results-zelda-target-set}
\end{table*}

Refer to Table \ref{tab:results-chatel-other-datasets} for a comparison of \approach~to the prompting method \textsc{ChatEL}\footnote{We report the numbers that are presented in the README of the repository (\url{https://github.com/yifding/In_Context_EL}), for which they used GPT-3.5.}, as well as the BLINK models, on the subset of datasets for which all three models report numbers, including MSNBC \cite{cucerzan-2007-large}, ACE2004 \cite{ratinov-etal-2011-local} and AQUAINT \cite{mine-2008-learning}.
As these are not part of the ZELDA splits, it is not guaranteed that all their target labels are included in our employed label set. In fact, we found that \textasciitilde15\% of their target labels are missing, mostly due to changes in article names. For example, one gold label in AQUAINT is the article "Dave\_Richardson", an article now redirecting to "David\_Richardson", which is a disambiguation page that links to, among others, two different cricket players with the same name. To ensure a fair evaluation, we add the missing labels (along with their verbalizations) to our pool. Keep in mind that while this ensures the inclusion of all labels, this may lead to a) incomplete verbalizations for outdated entity labels, and b) possibly multiple versions of the same entity, decreasing the likelihood of selecting the "correct" one.

On average over all datasets for which we could compare scores, \approach~ beats \textsc{ChatEL} and the BLINK Bi-Encoder, while on MSNBC and ACE2004 \textsc{ChatEL} performs better, possibly due to the discrepancy in the label sets. Note that the performance improvements after incorporating additional labels highlight \approach’s ability to adapt to unseen labels without requiring retraining.

\subsection{Results with Target Label Set}
\label{target-label-set}

In Table \ref{results-zelda-target-set}, we report results of our main models trained with the general label set on ZELDA, but restricting the label set to each respective target label set for inference. As expected, this simplification leads to significantly higher accuracy across all datasets. The findings indicate that training with a broader label set does not compromise performance on more specific label sets.

\end{document}